\documentclass[10pt,twocolumn,letterpaper]{article}

\usepackage[pagenumbers]{cvpr}

\usepackage{graphicx}
\usepackage{amsmath}
\usepackage{amssymb}
\usepackage{booktabs}

\usepackage{lib}

\usepackage{bm}
\usepackage[table]{xcolor}
\usepackage{stmaryrd}
\usepackage{multirow}

\newcommand{\parnobf}[1]{\vspace{1mm} \par \noindent {\bf {#1}}}

\usepackage[pagebackref,breaklinks,colorlinks]{hyperref}
\usepackage{epigraph}

\usepackage{color}
\definecolor{crimson}{rgb}{0.86, 0.08, 0.24}
\definecolor{gray}{rgb}{0.5,0.5,0.5}
\definecolor{green}{rgb}{0, 0.4, 0}
\definecolor{orange}{rgb}{1, 0.5, 0}
\definecolor{mahogany}{rgb}{0.75, 0.25, 0.0}
\definecolor{purple}{rgb}{0.6, 0, 0.6}
\definecolor{darkgreen}{rgb}{0, 0.4, 0}
\definecolor{frenchblue}{rgb}{0.0, 0.45, 0.73}
\definecolor{blue}{rgb}{0.0, 0.0, 0.65}
\definecolor{red}{rgb}{1,0,0}
\definecolor{yellow}{rgb}{1,1,0}
\definecolor{magenta}{rgb}{1,0,1}
\definecolor{pink}{rgb}{1,0.412,0.706}
\definecolor{newgreen}{rgb}{0, 0.6, 0.2}

\newlength\paramargin
\newlength\figmargin
\newlength\subfigmargin
\newlength\subsecmargin
\newlength\tabmargin
\newlength\eqmargin

\newlength\presecmargin
\newlength\secmargin

\usepackage[capitalize]{cleveref}
\crefname{section}{Sec.}{Secs.}
\Crefname{section}{Section}{Sections}
\Crefname{table}{Table}{Tables}
\crefname{table}{Tab.}{Tabs.}

\def\cvprPaperID{7031} 
\def\confName{CVPR}
\def\confYear{2023}

\begin{document}

\title{
SparseFusion: Distilling View-conditioned Diffusion for 3D Reconstruction
}

\author{Zhizhuo Zhou \qquad Shubham Tulsiani \\
Carnegie Mellon University \\
{\tt \small \{zhizhuo, shubhtuls\}@cmu.edu} 
\\ {\tt \small \href{https://sparsefusion.github.io/}{https://sparsefusion.github.io/}}
}

\twocolumn[{
\maketitle
\vspace{-3mm}
\renewcommand\twocolumn[1][]{#1}
\centering 
\vspace{-2.0mm}
\includegraphics[width=1.0\linewidth]{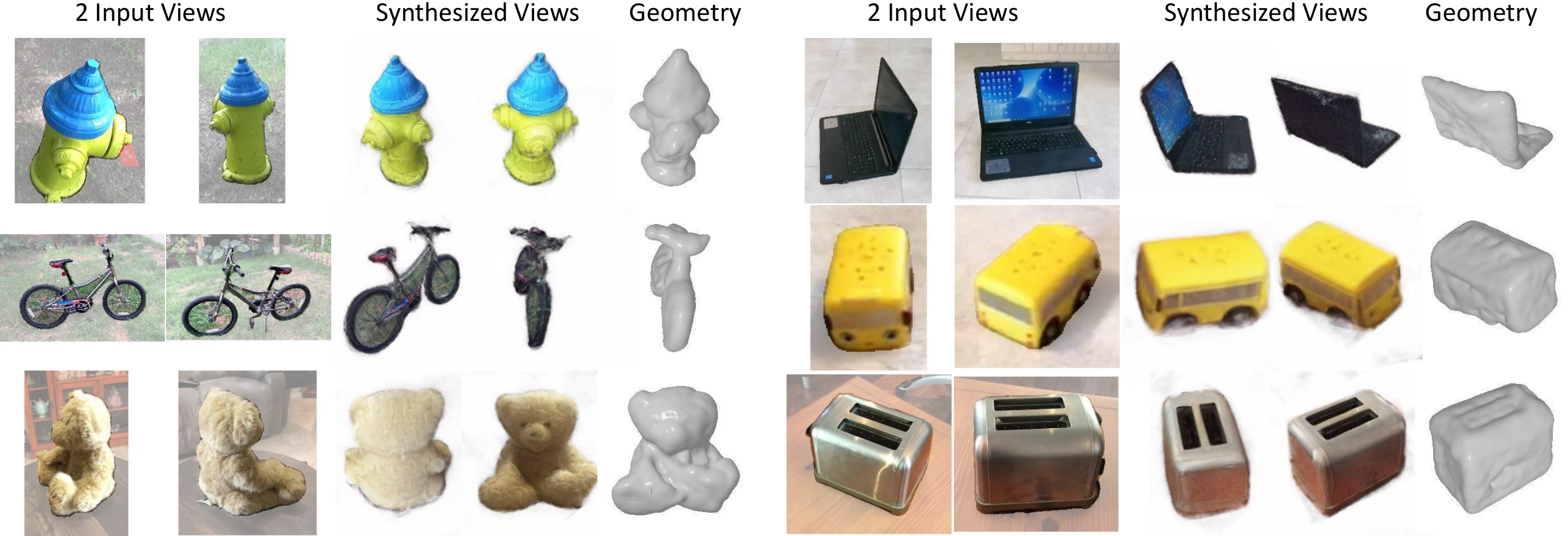}
\captionof{figure}{
    \textbf{Sparse-view Reconstruction.} We present SparseFusion, an approach for 3D reconstruction given a few (\eg just two) segmented input images with known relative pose. SparseFusion is able to generate a 3D consistent neural scene representation, enabling us to render novel views and extract the underlying geometry, while being able to generate detailed and plausible structures in uncertain or unobserved regions (\eg front of hydrant, teddy's face, back of laptop, or left side of toybus). 
    Please see project page for 360-degree visualizations.
} 
\figlabel{teaser}
\vspace{3mm}

}]


\begin{abstract}
\vspace{-3mm}
We propose SparseFusion, a sparse view 3D reconstruction approach that unifies recent advances in neural rendering and probabilistic image generation. Existing approaches typically build on neural rendering with re-projected features but fail to generate unseen regions or handle uncertainty under large viewpoint changes. Alternate methods treat this as a (probabilistic) 2D synthesis task, and while they can generate plausible 2D images, they do not infer a consistent underlying 3D. However, we find that this trade-off between 3D consistency and probabilistic image generation does not need to exist. In fact, we show that geometric consistency and generative inference can be complementary in a mode-seeking behavior. By distilling a 3D consistent scene representation from a view-conditioned latent diffusion model, we are able to recover a plausible 3D representation whose renderings are both accurate and realistic. We evaluate our approach across 51 categories in the CO3D dataset and show that it outperforms existing methods, in both distortion and perception metrics, for sparse-view novel view synthesis.

\end{abstract}
\vspace{-0.5em}
\section{Introduction}
\seclabel{introduction}

Consider the two images of the teddybear shown in \figref{teaser} and try to imagine the underlying 3D object. Relying on the direct visual evidence in these images, you can easily infer that the teddybear is white, has a large head, and has small arms. Even more remarkably, 
you can imagine beyond the directly visible to estimate a \emph{complete} 3D model of this object \eg forming a mental model of the teddy's face with (likely black) eyes even though these were not observed. In this work, we build a computational approach that can similarly predict 3D from just a few images -- by integrating visual measurements and priors via probabilistic modeling and then seeking likely 3D modes.

A growing number of recent works have studied the related tasks of \emph{sparse-view} 3D reconstruction and novel view synthesis, \ie inferring 3D representations and/or synthesizing novel views of an object given just a few (typically 2-3) images with known relative camera poses.  By leveraging data-driven priors, these approaches can learn to efficiently leverage  multi-view cues  and infer 3D from sparse views. However, they still yield blurry predictions under large viewpoint changes and cannot hallucinate plausible content in unobserved regions. This is because they do not account for the uncertainty in the outputs \eg the unobserved nose of a teddybear may be either red or black, but these methods, by reducing inference to independent pixel-wise or point-wise predictions, cannot model such variation.

In this  work, we propose to instead model the \emph{distribution} over the possible images  given observations from some context views and an arbitrary query viewpoint. Leveraging a geometrically-informed backbone that computes pixel-aligned features in the query view, our approach learns a (conditional) diffusion model that can then infer detailed plausible novel-view images. While this probabilistic image synthesis approach allows the generation of higher quality image outputs, it does not directly yield a 3D representation of underlying the object. In fact, the (independently) sampled outputs for each query view often do not even correspond to a consistent underlying 3D \eg if the nose of the teddybear is unobserved in context views, one sampled query view may paint it red, while another one black.

To obtain a consistent 3D representation, we propose a \emph{Diffusion Distillation} technique that `distills' the predicted  distributions into an instance-specific 3D representation. We note that the conditional diffusion model not only gives us the ability to sample novel-view images but also to (approximately) compute the likelihood of a generated one. Using this insight, we optimize an instance-specific (neural) 3D representation by maximizing the diffusion-based likelihood of its renderings. We show that this leads to a mode-seeking optimization that results in more accurate and realistic renderings, while also recovering a 3D-consistent representation of the underlying object. We demonstrate our approach on over 50 real-world categories from the CO3D dataset and show that our method allows recovering accurate 3D and novel views given as few as 2 images as input -- please see \figref{teaser} for sample results. 

\section{Related Work}
\seclabel{related_work}

\newcommand{\rot}[2][l]{\rotatebox[origin=#1]{90}{#2}} 
\definecolor{checkyes}{rgb}{0.7, 1.0, 0.7}
\definecolor{crossno}{rgb}{1.0, 0.7, 0.7}
\def \y {$\checkmark$\cellcolor{checkyes}}
\def \n {$\times$\cellcolor{crossno}}

\begin{table}[t]
\centering
\setlength{\tabcolsep}{3pt}
\def\arraystretch{1}
\resizebox{\linewidth}{!}{
\begin{tabular}{lcccc|cccc|ccc|c}
\toprule
 & \multicolumn{4}{c}{Single-instance}
 & \multicolumn{4}{c}{Re-projection}
 & \multicolumn{3}{c}{Latent}
 
 & Ours
 \\
 \cmidrule(lr){2-5}
 \cmidrule(lr){6-9}
 \cmidrule(lr){10-12}
 \cmidrule(lr){13-13}
 &
  \rot{NeRF~\cite{mildenhall2020nerf}} &
  \rot{RegNeRF~\cite{niemeyer2022regnerf}} &
  \rot{VolSDF~\cite{yariv2021volsdf}} &
  \rot{NeRS~\cite{zhang2021ners}} &

  \rot{IBRNet~\cite{wang2021ibrnet}} &
  \rot{PixelNeRF~\cite{yu2021pixelnerf}} &
  \rot{NerFormer~\cite{reizenstein2021common}} &
  \rot{GPNR~\cite{suhail2022generalizable}} &
  
  \rot{LFN~\cite{sitzmann2021light}} &
  \rot{SRT~\cite{sajjadi2022scene}} &
  \rot{ViewFormer~\cite{kulhanek2022viewformer}} &

    \rot{SparseFusion}\\
\cmidrule{2-13}
{1) Real data}      & \y & \y & \y & \y & \y & \y & \y & \y & \n & \y &  \y & \y \\
{2) Sparse-views}   & \n & \y & \y & \y & \n & \y & \y & \n & \y & \y &  \y & \y \\
{3) 3D consistent}  & \y & \y & \y & \y & \y & \y & \y & \y & \y & \y &  \n & \y \\
{4) Generalization}     & \n & \n & \n & \n & \y & \y & \y & \y & \y & \y &  \y & \y \\
{5) Generate unseen}     & \n & \n & \n & \n & \n & \n & \n & \n & \n & \n &  \y & \y \\
\bottomrule
\end{tabular}
}
\vspace{-.5em}
\caption{
    \textbf{Comparison with prior methods.}
    The rows indicate whether each method:
    1) has been demonstrated on real world data, 
    2) works with sparse (2-6) input views, 
    3) generates geometrically consistent views,
    4) generalizes to new scene instances,
    and 5) hallucinates unseen regions. 
    }
\vspace{-5mm}
\tablelabel{intro}
\label{tab:intro}

\end{table}

\vspace{-2mm}
\paragraph{Instance-specific Reconstruction from Multiple Views.} 

Leveraging Structure-from-Motion \cite{snavely2008modeling, schonberger2016structure} to recover camera viewpoints, early Multi-view-Stereo (MVS)~\cite{furukawa2009accurate,seitz2006comparison} methods could recover dense 3D outputs. Recent neural incarnations of these \cite{mildenhall2020nerf,yariv2021volsdf, yariv2020multiview} use volumetric rendering to learn a compact neural scene representation. Follow up works \cite{mueller2022instant, Chen2022tensorf, fridovich2022plenoxels} seek to make the training and rendering orders of magnitudes faster.
However, these methods require many input views, making them impractical for real world applications. While some works \cite{zhang2021ners, goel2022differentiable, niemeyer2022regnerf} seek to reduce the input views required, they still do not make predictions for unseen regions.

\vspace{-3mm}
\paragraph{Single-view 3D Reconstruction.} 
The ability to predict 3D geometry (and appearance) beyond the visible is a key goal for single-view 3D prediction methods. While these approaches have pursued prediction of different 3D representations \eg volumetric~\cite{girdhar2016learning,choy20163d,tulsiani2017multi,henzler2019escaping,ye2021shelf}, mesh-based~\cite{gkioxari2019mesh,kanazawa2018learning}, or neural implicit~\cite{mescheder2019occupancy,vasudev2022ss3d,lin2020sdfsrn} 3D, the use of a single input image fundamentally limits the details that can be predicted. Moreover, these methods do not prioritize view synthesis as a goal. While our approach similarly learns data driven inference, we aim for a more detailed reconstruction and high quality novel-view renderings.

\begin{figure*}[t!]
    \centering
    \includegraphics[width=1.0\linewidth]{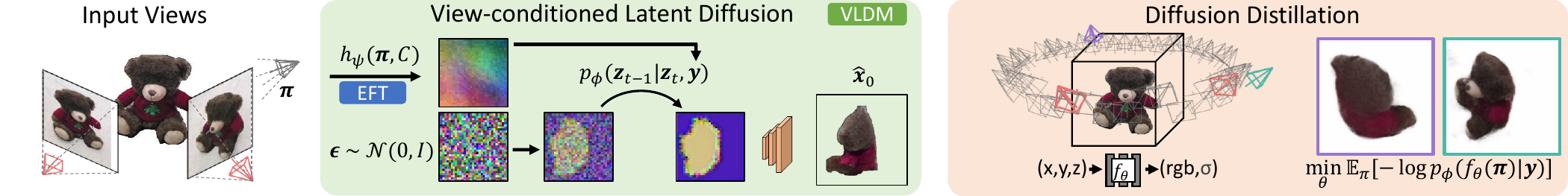}
    \caption{
    \textbf{Overview of SparseFusion.} SparseFusion comprises of two core components: a view-conditioned latent diffusion model (VLDM) and a diffusion distillation process that optimizes an Instant NGP \cite{mueller2022instant, torch-ngp}. We use VLDM to model $p(\bm{x}|\bm{\pi}, C).$}
    \vspace \figmargin
    \figlabel{overview}
\end{figure*} 
\vspace{-2mm}
\paragraph{Generalizable View Synthesis from Fewer Views.} 
Novel view synthesis (NVS), while similar to reconstruction, has slightly different roots. Earlier works~\cite{zhou2016view, tatarchenko2016multi} frame NVS as a 2D problem, using deep networks to make predictions from global encodings. Recent approaches combine deep networks with various rendering formulations~\cite{sitzmann2019scene, sitzmann2021light, sajjadi2022scene}. Strong performing approaches often leverage re-projected features from input views with volumetric rendering~\cite{yu2021pixelnerf, trevithick2021grf, reizenstein2021common} or image based rendering~\cite{wang2021ibrnet, suhail2022generalizable, Cao2022FWD}. While feature re-projection methods are 3D consistent, they regress to the mean and fail to produce perceptually sharp outputs. Another line of work~\cite{rombach2021geometry, kulhanek2022viewformer} revisits NVS as a probabilistic 2D generation task, using newer generative backbones to offer better perceptual quality at the cost of larger distortion and 3D consistency. See \tableref{intro} for a comparison of our method against existing approaches. 

\vspace{-2mm}
\paragraph{Diffusion Models.} 
Several works extend upon denoising diffusion models~\cite{ho2020denoising, song2020denoising} to achieve impressive applications, such as generating images from text \cite{ramesh10hierarchical, saharia2022photorealistic} and placing foreground objects in different backgrounds \cite{ruiz2022dreambooth}. In this work, we leverage this class of models for (probabilistic) novel view synthesis while using geometry-aware features as conditioning. Inspired by the impressive results in DreamFusion ~\cite{poole2022dreamfusion} which optimized 3D scenes using text-conditioned diffusion models, we propose a view-conditioned diffusion distillation mechanism to similarly extract 3D modes in the sparse view reconstruction task.

\paragraph{Concurrent Works.}
Several concurrent works also leverage diffusion models for 3D reconstruction and view synthesis.
3DiM \cite{watson2022novel} proposes a 2D diffusion approach for image-conditioned novel view synthesis, but does not infer a 3D representation like our approach. Closer to our work, Deng \etal \cite{deng2022nerdi} uses (pre-trained) 2D diffusion models as guidance for single-view 3D, but obtain coarser reconstructions in this more challenging setting. While we leverage a 2D diffusion model for optimizing 3D,  RenderDiffusion \cite{anciukevivcius2022renderdiffusion} learns a diffusion model in 3D space. Concurrently to DreamFusion~\cite{poole2022dreamfusion}, which inspired our distillation objective, Wang \etal~\cite{wang2022score} provide a  different mathematical intuition for a similar objective. 

\section{Background: Denoising Diffusion}
\seclabel{preliminaries}

Our method adopts and optimizes through denoising diffusion models~\cite{ho2020denoising}, and here we give a brief summary of the key formulations used, and refer the reader to the appendix for further details. 

\vspace{-2mm}
\paragraph{Training Objective.} One can learn denoising diffusion models by optimizing a variational lower bound on the log-likelihood of the observed data. Conveniently, this reduces to a training framework where one adds (time-dependent) noise to a data point $\bm{x}_0$, and then trains a network $\bm{\epsilon}_\phi$ to predict this noise given the noisy data point $\bm{x}_t$.
\begin{equation} \eqlabel{diffusion}
\begin{split}
\mathcal{L}_{DM} = \mathbb{E}_{{\bm{x}_0}, \bm{\epsilon}, t}\left[w_t~||\bm{\epsilon} - \bm{\epsilon}_\phi(\bm{x}_t, t)||^2\right]
\\ \text{where}~\bm{x}_t = \sqrt{\bar{\alpha_t}} \bm{x}_0 + \sqrt{1-\bar{\alpha_t}} \bm{\epsilon}; ~~ \bm{\epsilon} \sim \mathcal{N}(0,1)
\end{split}
\end{equation}
Here, $\bar{\alpha}_t$ is a scheduling hyper-parameter, and the weights $w_t$ depend on this learning schedule, but are often set to 1 to simplify the  objective.

\vspace{-4mm}
\paragraph{Interpretation as Reconstruction Error.} The above noise prediction objective, which represents a  bound on the log likelihood, can also be viewed as a reconstruction error. Concretely, given a noisy $\bm{x}_t$, the network prediction $\bm{\epsilon}_\phi(\bm{x}_t, t)$ can  be interpreted as yielding a  reconstruction for the original input, where the learning objective can be rewritten as a reconstruction error:
\begin{equation} \eqlabel{xstart}
\hat{\bm{x}}_{0,t} = \frac{\bm{x}_t - \sqrt{1-\bar{\alpha}_t} \bm{\epsilon}_\phi(\bm{x}_t, t)}{\sqrt{\bar{\alpha}_t}}
\end{equation}
\begin{equation} \eqlabel{diffusionimage}
\mathcal{L}_{DM} = \mathbb{E}_{{\bm{x}_0}, \bm{\epsilon}, t}\left[w'_t~||\hat{\bm{x}}_{0,t} - \bm{x}_0||^2\right]
\end{equation}

While the above summary focused on unconditional diffusion models, they can be easily extended to infer conditional distributions $p(\bm{x}|\bm{y})$ by additionally using $\bm{y}$ as an input for the noise prediction network $\bm{\epsilon}_\phi$. 

\section{Approach}
\label{sec:approach}

\begin{figure}[t!]
    \centering
    \includegraphics[width=1.0\linewidth]{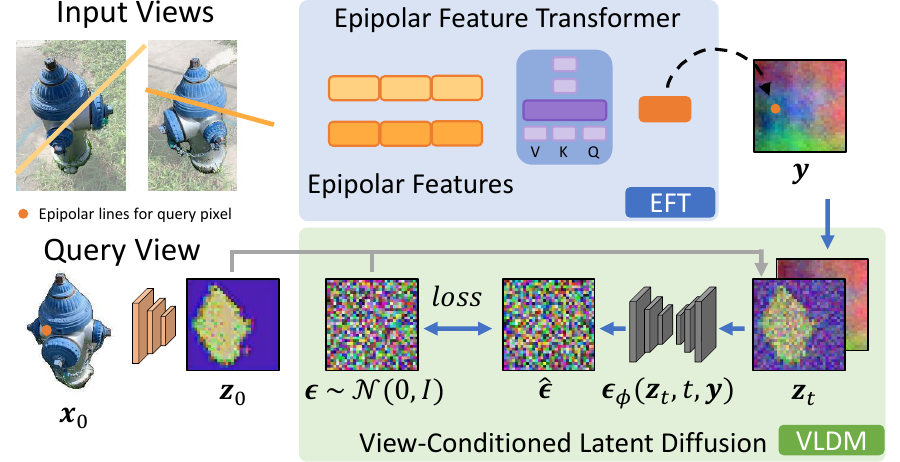}
    \caption{
    \textbf{View-conditioned Diffusion.} We show a diagram of our view-conditioned latent diffusion model. VLDM is conditioned on features $\bm{y}$, which is predicted by EFT.}
    \vspace \figmargin
    \figlabel{vldm}
\end{figure}

Given sparse-view observations of an object (typically 2-3 images with masked foreground) with known camera viewpoints, our approach aims to infer a (3D) representation capable of synthesizing novel views while also capturing the geometric structure. However, as aspects of the object may be unobserved and its geometry difficult to precisely infer, direct prediction of 3D or novel views leads to implausibly blurry outputs in regions of uncertainty.

\begin{figure*}[t!]
    \centering
    \includegraphics[width=1.0\linewidth]{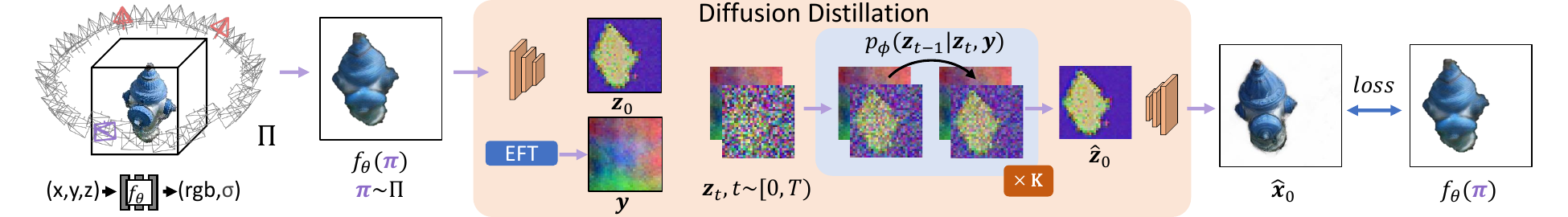}
    \caption{
    \textbf{Diffusion Distillation Diagram.} We optimize the parameters $\theta$ of an Instant NGP network such that rendered images $f_\theta(\bm{\pi})$ from $\bm{\pi} \sim \Pi$ are similar to VLDM predictions $\hat{\bm{x}}_0$, effectively seeking a mode in $p_\phi(\bm{x}|\bm{\pi}, C)$.}
    \vspace \figmargin
    \figlabel{distillation}
\end{figure*}

To enable plausible and 3D-consistent predictions, we instead take a two step approach as outlined in \figref{overview}. First, we learn a probabilistic view-synthesis model that, using geometry-guided diffusion, can model the \emph{distribution} of images from query views given the sparse-view context (\secref{probview}).  While this allows the generation of detailed and diverse outputs, the obtained renderings lack 3D consistency. To extract a 3D representation, we propose a 3D neural distillation process that `distills' the predicted view distributions into a consistent 3D mode (\secref{diffdist}).

\subsection{Geometry-guided  Probabilistic View Synthesis}
\seclabel{probview}

Given a target view pose $\bm{\pi}$ along with a set of reference images and their relative poses $C \equiv {(\bm{x}_m, \bm{\pi}_m})$, we want to model the conditional distribution $p(\bm{x}|\bm{\pi}, C)$, from which we can synthesize an image $\bm{\hat{x}}$.
We illustrate our approach to modeling this distribution in  \figref{vldm}. First, we use an epipolar feature transformer (EFT) inspired by \cite{suhail2022generalizable} as feature extractor to obtain a low resolution feature grid $\bm{y}$ in the view space of $\bm{\pi}$ given the context $C$. In conjunction, we train a view-conditioned latent diffusion model (VLDM) that models the distribution over novel-view images condition on these geometry-aware features.

\vspace{-.5em}

\subsubsection{Epipolar Feature Transformer}
\vspace{-2mm}
We build upon GPNR \cite{suhail2022generalizable} to extract features from context $C$. GPNR learns a feedforward network, $g_\psi(\bm{r}, C)$, that predicts color given a query ray $\bm{r}$ by extracting features along its epipolar lines in all context images and aggregating them with transformers. We make several modifications to GPNR to suit our needs. First, we replace the patch projection layer with a ResNet18 \cite{he2016deep} convolutional encoder as we found the lightweight patch encodings, while suitable for small baseline view synthesis, are not robust under the sparse-view setting. Furthermore, we modify the last layer to predict both an RGB value and a feature vector. We denote the RGB branch as $g_\psi$ and the feature branch as $h_\psi$. We refer to our modified epipolar patch-based feature transformer as \textbf{EFT} and present its color branch as a strong baseline. 

We train the color branch of the EFT to minimize a simple reconstruction loss in \eqref{gpnr}, where $\bm{r}$ is a query ray sampled from $\bm{\pi}$, $C$ is the set of reference images and their relative poses, and $I(\bm{r})$ is the ground truth pixel value.
\begin{equation} \eqlabel{gpnr}
\mathcal{L}_{EFT} = \sum_{\bm{r}\in R(\bm{\pi})} ||g_\psi(\bm{r}, C) - I(\bm{r})||^2
\end{equation}

\vspace{-.8em}

\begin{figure}[t!]
\centering 
\vspace{-0.5mm}
\includegraphics[width=1.0\linewidth]{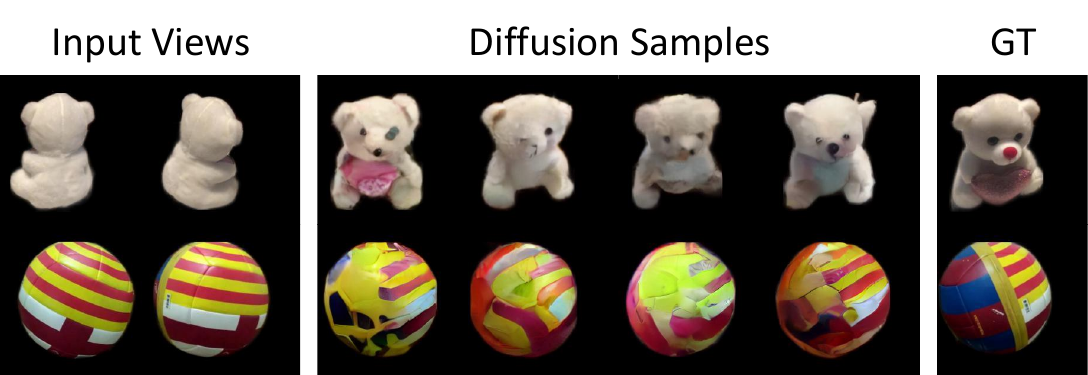}
\captionof{figure}{
    \textbf{Diffusion Samples.} Given the same input features, the reverse sampling process of diffusion model results in different predictions for unseen regions. 
} 
\label{fig:diffusion-samples}
\figlabel{diffusion-samples}
\vspace{-5mm}
\end{figure}
\begin{figure}[t!]
    \centering
    \includegraphics[width=0.95\linewidth]{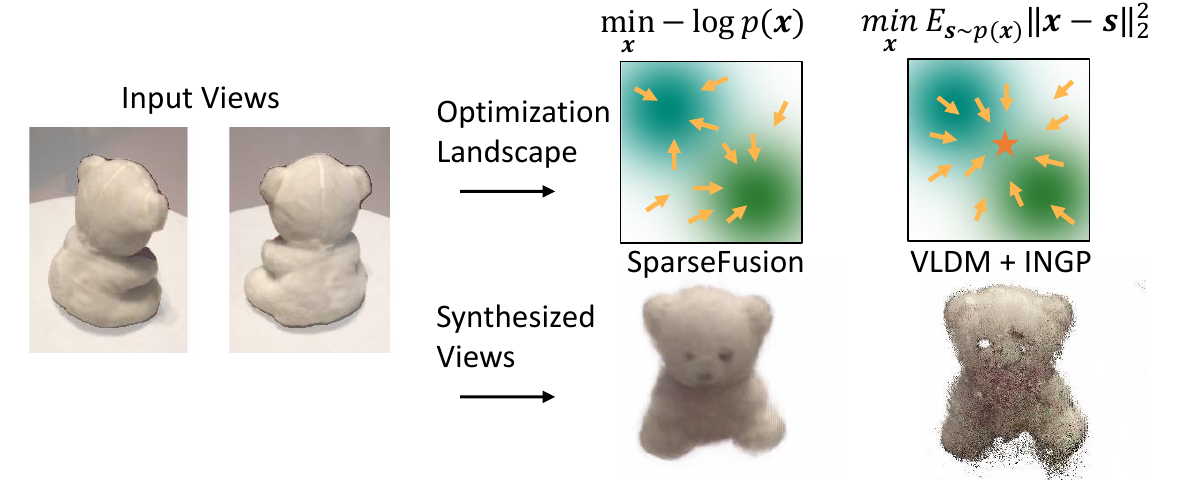}
    \vspace{-1em} 
    \caption{
    \textbf{Mode Seeking Visualization.} We show qualitative comparison between a mode-seeking (SparseFusion) and a mean-seeking (VLDM+INGP) objective. }
    \vspace \figmargin
    \figlabel{mode}
\end{figure} 

\subsubsection{View-conditioned Latent Diffusion Model}

While EFT can directly predict novel views, the pixelwise prediction mechanism does not allow it to model the underlying probability distribution, thus resulting in blurry mean-seeking predictions under uncertainty. To model the distribution over plausible images, we train a view-conditioned diffusion model to estimate $p(\bm{x}|\bm{\pi}, C)$ while using EFT as a geometric feature extractor. Instead of directly modeling the distribution in pixel space, we find it computationally efficient to do so in a lower-resolution latent space $\bm{z} = \mathcal{E}(\bm{x})$, which can be decoded back to an image as $\bm{x} = \mathcal{D}(\bm{z})$.
Please see the appendix for details.

Given target view $\bm{\pi}$ and a set of input images $C$, we extract a 32 by 32 feature grid $\bm{y} = h_\psi(\bm{\pi}, C)$ using the EFT backbone. We train our VLDM to recover ground truth image latent $\bm{z_0}$ conditioned on $\bm{y}$. Following diffusion model training conventions \cite{ho2020denoising, song2020denoising,ramesh10hierarchical}, we optimize a simplified variational lower bound in \eqref{latentdiffusion}. 
\vspace{-2mm}
\begin{equation} \eqlabel{latentdiffusion}
\mathcal{L}_{VLDM} = \mathbb{E}_{\bm{z}, \bm{\epsilon} ~ \mathcal{N}(0,1), t, \bm{y}}\left[||\bm{\epsilon} - \bm{\epsilon}_\phi(\bm{z_t}, t, \bm{y})||^2\right]
\end{equation}
\figref{vldm} shows a diagram of the training setup. Our VLDM model allows us to approximate $p(\bm{x}|\bm{\pi}, C)$, and enables drawing multiple sample predictions. In \figref{diffusion-samples}, we see variations in VLDM predictions. Nevertheless, all predictions are plausible explanations for the target view given that majority of it is unseen.

\subsection{Extracting 3D Modes via Diffusion Distillation}
\seclabel{diffdist}
While the proposed VLDM gives us the ability to hallucinate unseen regions  and make realistic predictions under uncertainty, it does not output a 3D representation. In fact, as it models the distribution over images, the views sampled from the VLDM do not (and should not!) necessarily correspond to a single  underlying 3D interpretation. How can we then obtain an output 3D representation while preserving the high-quality of renderings? 

\vspace{-2mm}
\paragraph{3D Inference as Neural Mode Seeking.}
Our key insight is that the VLDM model not only allows us to sample plausible novel views, but the modeled distribution also gives us a mechanism to approximate the likelihood of a generated novel view. Building on this insight, we propose to distill the VLDM predictions to obtain an instance-specific 3D neural scene representation $f_{\theta}$, such as NeRF \cite{mildenhall2020nerf} or Instant NGP (INGP) \cite{mueller2022instant}. Intuitively, we want to arrive at a solution for $f_{\theta}$ such that its renderings $\bm{x} \equiv f_{\theta}(\bm{\pi})$ from arbitrary viewpoints $\bm{\pi}$ are likely under the conditional distribution modeled by the VLDM $ p_\phi(\bm{x} | \bm{\pi}, C)$:

\begin{equation} \eqlabel{distillationsetup}
\underset{\theta}{\min} \; \mathbb{E}_{\bm{\pi} \sim \Pi} \; -\log \, p_\phi(f_\theta(\bm{\pi}) | \bm{\pi}, C)
\end{equation}
where we minimize the negative log-likelihood for images rendered with $f_\theta$ over cameras sampled from a prior camera distribution $\Pi$ (constructed by assuming a circular camera trajectory and that all cameras look at a common center). We term this process as `neural mode seeking' as it encourages a representation which maximizes likelihood as opposed to minimizing distance to samples (mean seeking).

\vspace{-2mm}
\paragraph{Neural Mode Seeking via Diffusion Distillation.}
Given a learned diffusion model, the reconstruction objective (\eqref{diffusionimage}) yields a bound on the log-likelihood of a data point $\bf{x}$.
This approximation yields a simple mechanism for computing the likelihood of a (rendered) image $f_{\theta}(\bm{\pi})$ to be used in the mode-seeking optimization (\eqref{distillationsetup}):
\begin{equation} \eqlabel{modesampling}
-\log p_{\phi}(\bm{x}_0) \approx \mathbb{E}_{\bm{\epsilon}, t}\left[w_t||\bm{z}_0 - \hat{\bm{z}}_{0,t}||^2\right] + C
\end{equation}
where $\bm{z}_0 = \mathcal{E}(f_\theta(\bm{\pi}))$ is the latent of the rendered image, $t \sim (0,  T]$, and $\hat{\bm{z}}_{0,t}$ is the predicted latent  (analogous to $\hat{\bm{x}}_{0,t}$ in \eqref{xstart}). 
Intuitively, this objective implies that if, after adding noise to obtain $\bm{z}_t$ from $\bm{z}_0$, the denoising diffusion model predicts $\hat{\bm{z}}_0$ close to the original input, one has reached a mode under $p_{\phi}(\bm{z})$. We visualize the behavior of mode seeking versus mean seeking in \figref{mode}.

\vspace{-2mm}
\paragraph{Multi-step Denoising and Image-space Reconstruction.}
In practice, we make three modifications to the single-step objective in \eqref{modesampling} for better performance: 1) taking loss in pixel space instead of latent space \ie using $\bm{x}_0$ instead of $\bm{z}_0$, 2) using perceptual distance \cite{zhang2018unreasonable} in addition to the pixelwise distance, and 3) performing multi-step denoising. Instead of directly predicting $\hat{\bm{z}}_{0,t}$, we adaptively use multiple time-steps (up to 50 steps) $\mathcal{T} = (t_1, \cdots, t_k, t)$, and successively predict $\hat{\bm{z}}_{t_{k-1},t_{k}}$ (via \cite{liu2022pseudo}) \ie predict a denoised estimate for time $t_{k-1}$ given a sample from time $t_k$. We denote this reconstruction as $\hat{\bm{z}}_{0,\mathcal{T}}$ to highlight the multiple-step reconstruction.
 We express our final objective for optimizing for neural mode seeking with view-conditioned diffusion models as: 
\begin{equation} \label{eq:diffusiondistillation}
\mathcal{L} = \mathbb{E}_{\bm{\pi}, \bm{\epsilon}, t}\left[w_t||f_\theta(\bm{\pi}) - \hat{\bm{x}}_{0,{\mathcal{T}}}||^2 + \mathcal{L}_{Perp}(f_\theta(\bm{\pi}), \hat{\bm{x}}_{0,{\mathcal{T}}})\right]
\end{equation}
where $\hat{\bm{x}}_{0,{\mathcal{T}}} = \mathcal{D}(\hat{\bm{z}}_{0,{\mathcal{T}}})$, and $\hat{\bm{z}}_{0,{\mathcal{T}}}$ is the multi-step reconstruction from $\bm{z}_t$ -- which is obtained by adding noise to $\bm{z}_0 = \mathcal{E}(f_\theta(\bm{\pi}))$. While $\hat{z}$ in the above objective does (indirectly) depend on the neural representation $f_{\theta}$, we follow ~\cite{poole2022dreamfusion} in ignoring this dependence when computing parameter gradients (see ~\cite{wang2022score} for a justification). We outline the multi-step denoising diffusion distillation in \figref{distillation}.

\section{Experiments}
\label{sec:experiments}

\begin{figure*}[t!]
    \centering
    \includegraphics[width=1.0\linewidth]{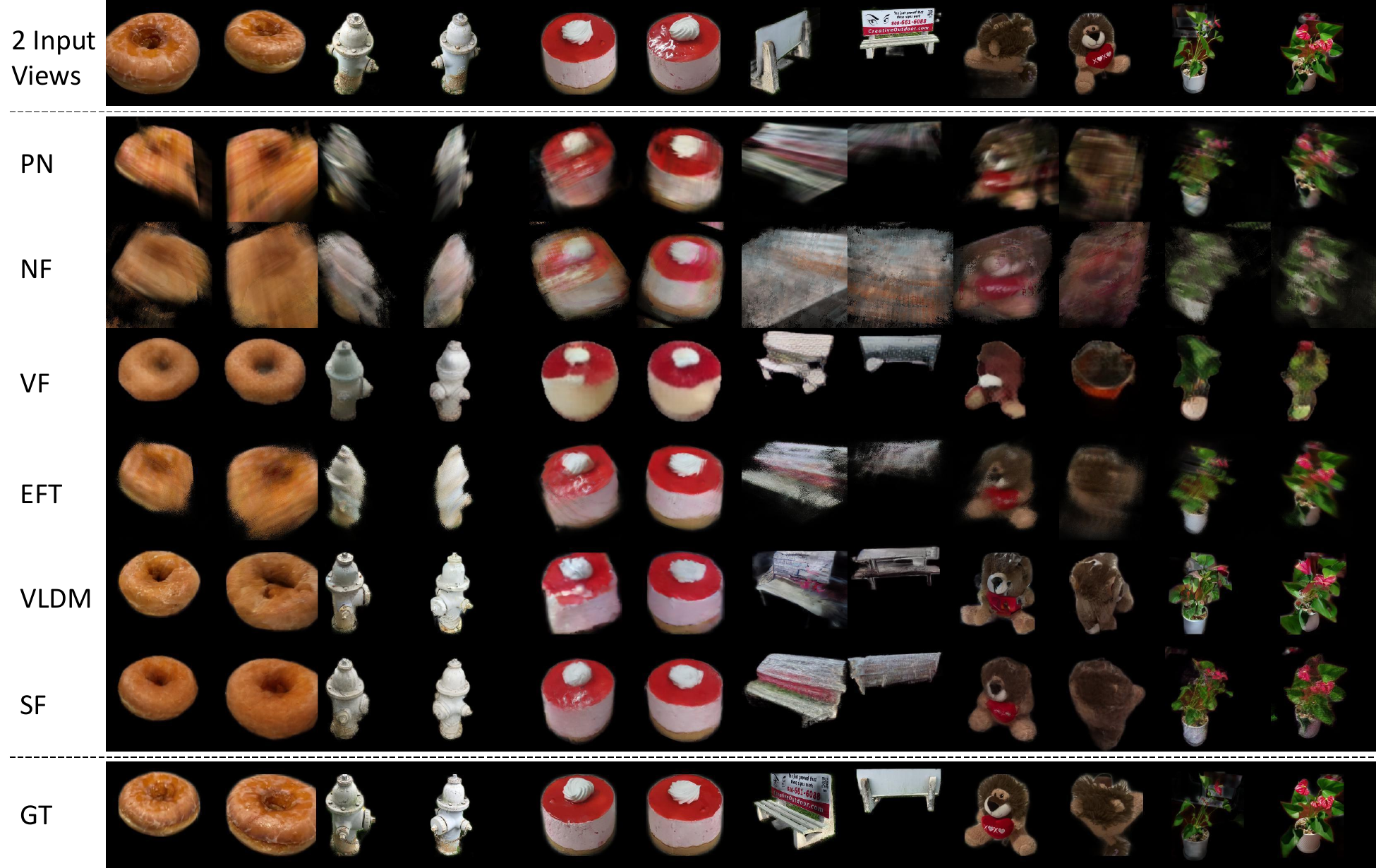}
    \caption{
    \textbf{View Synthesis Qualitative Results.} We show view synthesis results with 2 input views on donut, hydrant, cake, bench, teddybear, and plant categories. We visualize 2 novel views per instance with PixelNeRF (PN), NerFormer (NF), ViewFormer (VF), EFT, VLDM, and finally, SparseFusion (SF). Corresponding numbers can be found in \tableref{co3d-10d}.}
    \vspace \figmargin
    \figlabel{comparison-large}
\end{figure*} 

We demonstrate our approach on a challenging real world multi-view dataset CO3Dv2 \cite{reizenstein2021common}, across 51 diverse categories. First, we compare SparseFusion against prior works, highlighting the benefit of our approach in sparse view settings.  Then, we show the importance of diffusion distillation and its probabilistic mode-seeking formulation. 

\subsection{Experimental Setup}
\paragraph{Dataset.} We perform experiments on CO3Dv2 \cite{reizenstein2021common}, a multi-view dataset of real world objects annotated with relative camera poses and foreground masks. We use the specified \emph{fewview-train} and \emph{fewview-dev} splits for  training and evaluation. Since SparseFusion optimizes an instance-specific Instant NGP, it is computationally prohibitive to evaluate on all evaluation scenes. Instead, we perform most experiments on a \emph{\textbf{core subset}} of 10 categories proposed by \cite{reizenstein2021common}, evaluating 10 scenes per category. Furthermore, we demonstrate that SparseFusion extends to diverse categories by evaluating 5 scenes per category across 51 categories. 


\definecolor{first}{rgb}{1.0, .83, 0.3}
\definecolor{second}{rgb}{1.0, 0.93, 0.7}
\def \first {\cellcolor{first}}
\def \second {\cellcolor{second}}
\def \third {}

\begin{table*}[t!]
\footnotesize
\begin{center}
\vspace{1em}
\caption{\textbf{Detailed View Synthesis Benchmark.} We show 2-view category-specific metrics on 10 CO3D categories from the \emph{core subset}. We show PSNR $\uparrow$ and LPIPS $\downarrow$ averaged across 10 scenes per category.}
\vspace{-1em}
\tablelabel{co3d-10d}
\setlength{\tabcolsep}{2pt}
\resizebox{\linewidth}{!}{
\begin{tabular}{l cc cc cc cc cc cc cc cc cc cc}
\toprule

& \multicolumn{2}{c}{Donut} 
& \multicolumn{2}{c}{Apple}
& \multicolumn{2}{c}{Hydrant} 
& \multicolumn{2}{c}{Vase} 
& \multicolumn{2}{c}{Cake} 
& \multicolumn{2}{c}{Ball} 
& \multicolumn{2}{c}{Bench} 
& \multicolumn{2}{c}{Suitcase} 
& \multicolumn{2}{c}{Teddybear} 
& \multicolumn{2}{c}{Plant}\\

\cmidrule(r){2-3} \cmidrule(r){4-5} \cmidrule(r){6-7} \cmidrule(r){8-9} \cmidrule(r){10-11}
\cmidrule(r){12-13} \cmidrule(r){14-15} \cmidrule(r){16-17} \cmidrule(r){18-19} \cmidrule(r){20-21}

& PSNR  & LPIPS 
& PSNR  & LPIPS
& PSNR  & LPIPS
& PSNR  & LPIPS
& PSNR  & LPIPS
& PSNR  & LPIPS
& PSNR  & LPIPS
& PSNR  & LPIPS
& PSNR  & LPIPS
& PSNR  & LPIPS
\\

\midrule
PixelNeRF \cite{yu2021pixelnerf}       
& \third 20.9 & 0.30 & 20.0 & 0.35 & 19.0 & 0.27 & \second 21.3 & 0.26 & 18.3 & 0.37
& 18.5 & 0.36 & \second 17.7 & 0.35 & \third 21.7 & 0.30 & \third 18.5 & 0.35 & \third 19.3 & 0.36\\

NerFormer \cite{reizenstein2021common}      
& 20.3 & 0.34 & 19.5 & 0.33 & 18.2 & 0.30 & 17.7 & 0.34 & 16.9 & 0.44
& 16.8 & 0.35 & 15.9 & 0.44 & 20.0 & 0.39 & 15.8 & 0.43 & 17.8 & 0.45\\

ViewFormer\textsuperscript{\ref{fn:one}} \cite{kulhanek2022viewformer}      
& 19.3 & \third 0.29 & 20.1 & \third 0.26 & 17.5 & \third 0.22 & 20.4 & \third 0.21 & 17.3 & \third 0.33
& 18.3 & 0.31 & 16.4 & \third 0.30 & 21.0 & \third 0.26 & 15.5 & 0.32 & 17.8 & \third 0.31\\

EFT                   
& \second 21.5 & 0.31 & \second 22.0 & 0.29 & \second 21.6 & 0.22 & 21.1 & 0.25 & \second 19.9 & 0.33
& \second 21.4 & 0.29 & \first \textbf{17.8} & 0.34 & \first \textbf{23.0} & 0.26 & \second 19.8 & 0.30 & \first \textbf{20.4} & 0.31\\

VLDM                     
& 20.1 & \second 0.25 & 21.3 & \second 0.22 & 20.1 & \second 0.18 & 20.2 & \second 0.20 & 18.9 & \second 0.30
& 20.3 & \second 0.25 & 16.6 & \second 0.29 & 21.3 & \second 0.23 & 17.9 & \second 0.27 & 18.9 & \second 0.27\\

SparseFusion            
& \first \textbf{22.8} & \first \textbf{0.22} & \first \textbf{22.8} & \first \textbf{0.20} & \first \textbf{22.3} & \first \textbf{0.16} & \first \textbf{22.8} & \first \textbf{0.18} & \first \textbf{20.8} & \first \textbf{0.28}
& \first \textbf{22.4} & \first \textbf{0.22} & \third 16.7 & \first \textbf{0.28} & \second 22.2 & \first \textbf{0.22} & \first \textbf{20.6} & \first \textbf{0.24} & \second 20.0 & \first \textbf{0.25}\\

\bottomrule
\vspace{-3.5em}
\end{tabular}
}

\end{center}

\end{table*}
\vspace{-2mm}
\paragraph{Baselines.}
We compare SparseFusion against current state-of-the-art methods. We first compare against \emph{PixelNeRF} \cite{yu2021pixelnerf}, a feature re-projection method. We adapt \emph{PixelNeRF} to CO3Dv2 dataset and train category-specific models on the 10 categories of the \emph{core subset}, each for 300k steps. We also compare against \emph{NerFormer} \cite{reizenstein2021common}, another feature re-projection method. We use category-specific models provided by the authors for all 51 categories. Moreover, we compare against \emph{ViewFormer\footnote{\label{fn:one}Only category-agnostic CO3Dv1 weights are  compatible with our evaluation. We use the 10-category weights for our \emph{core subset} experiments and all-category weights for our all category experiments. Despite this difference, the comparative results of ViewFormer against our baselines are consistent with the comparisons reported in their original paper.}} \cite{kulhanek2022viewformer}, an autoregressive image generation method, using models provided by the authors. Lastly, we present components of SparseFusion, EFT and VLDM, as strong baselines.


\begin{table*}
    \centering
    \begin{minipage}{0.62\textwidth}

\footnotesize
\caption{\textbf{View Synthesis on 10 Categories.} We benchmark view synthesis results averaged across 10 categories with 2, 3, and 6 input views.  }
\label{tab:co3d-10}
\vspace{-.8em}
\setlength{\tabcolsep}{2pt}
\resizebox{\linewidth}{!}{
\begin{tabular}{l ccc ccc ccc}
\toprule
& \multicolumn{3}{c}{2 Views} & \multicolumn{3}{c}{3 Views} & \multicolumn{3}{c}{6 Views}\\
\cmidrule(r){2-4} \cmidrule(r){5-7} \cmidrule(r){8-10}
& PSNR $\shortuparrow$ & SSIM $\shortuparrow$ & LPIPS $\shortdownarrow$  & PSNR $\shortuparrow$ & SSIM $\shortuparrow$ & LPIPS $\shortdownarrow$ & PSNR $\shortuparrow$ & SSIM $\shortuparrow$ & LPIPS $\shortdownarrow$ \\
\midrule
PixelNeRF \cite{yu2021pixelnerf}       

& 19.52 & 0.667 & 0.327 & 20.67 & 0.712 & 0.293 & 22.47 & 0.776 & 0.241  \\

NerFormer \cite{reizenstein2021common}       

& 17.88 & 0.598 & 0.382 & 18.54 & 0.618 & 0.367 & 19.99 & 0.661 &  0.332 \\

ViewFormer\textsuperscript{\ref{fn:one}} \cite{kulhanek2022viewformer}      

& 18.37 & 0.697 & 0.282 & 18.91 & 0.704 & 0.275 & 19.72 & 0.717 & 0.266  \\

EFT                   

& \second 20.85 & 0.680 & 0.289 & \first \textbf{22.71} & \second 0.747 & 0.262 & \first \textbf{24.57} & \first \textbf{0.804} & 0.210  \\

VLDM                     

& 19.55 & \second 0.711 & \second 0.247 & 20.85 & 0.737 & \second 0.225 & 22.35 & 0.768 & \second 0.201  \\

SparseFusion            

& \first \textbf{21.34} & \first \textbf{0.752} & \first \textbf{0.225} & \second 22.35 & \first \textbf{0.766} & \first \textbf{0.216} & \second 23.74 & \second 0.791 & \first \textbf{0.200}  \\

\bottomrule
\vspace{-3em}
\end{tabular}
}

    \end{minipage}\hfill
    \begin{minipage}{0.36\textwidth}
        \centering

\vspace{-.45em}
\footnotesize
\caption{\textbf{View Synthesis on 51 Categories.} We benchmark on all CO3D categories. }
\label{tab:co3d-50}
\vspace{-.7em}
\resizebox{21.6em}{!}{
\begin{tabular}{l ccc}
\toprule
& \multicolumn{3}{c}{2 Views} \\
\cmidrule(r){2-4}
& PSNR $\shortuparrow$ & SSIM $\shortuparrow$ & LPIPS $\shortdownarrow$ \\
\midrule
\\
NerFormer \cite{reizenstein2021common}           

& 18.44 & 0.614 & 0.365   \\

ViewFormer\textsuperscript{\ref{fn:one}} \cite{kulhanek2022viewformer}         

& 18.91 & 0.718 & 0.265  \\

EFT                         

& \first \textbf{21.44} & 0.719 & 0.281 \\

VLDM                       

& 19.85 & \second 0.732 & \second 0.229   \\

SparseFusion                

& \second 21.20 & \first \textbf{0.756} & \first \textbf{0.223}   \\

\bottomrule
\vspace{-3.5em}
\end{tabular}
}

    \end{minipage}
\vspace{.7em}
\end{table*}

\vspace{-2mm}
\paragraph{Metrics.}
We report standard image metrics PSNR, SSIM, and LPIPS \cite{zhang2018unreasonable}. We recognize that no metric is perfect for ambiguous cases of novel view synthesis; PSNR derives from pixelwise MSE and favors mean color prediction while SSIM and LPIPS favor perceptual agreement.

\vspace{-2mm}
\paragraph{Implementation Details.}
For EFT, we use a ResNet18 \cite{he2016deep} backbone and three groups of transformer encoders with 4 layers each. We use 256 hidden dimensions for all layers. For VLDM, we freeze the VAE from \cite{rombach2022high} that encodes 256x256 images to 32x32 latents with channel dimension of 4. We construct a 400M parameter denosing UNet similar to \cite{ronneberger2015u, saharia2022photorealistic} for probabilistic modeling.
We jointly train category-specific EFT and VLDM models, using \eqref{gpnr} and  \eqref{latentdiffusion}, across all categories in CO3Dv2. We use a batch size of 2 and train for 100K iterations.

For diffusion distillation, we use a PyTorch implementation of Instant NGP \cite{torch-ngp, mueller2022instant}. Due to memory constraints, we render images at 128x128 and upsample to 256x256 before performing diffusion distillation. For each instance, we optimize Instant NGP for 3,000 steps. During the first 1,000 steps, we optimize rendering loss on input images and predicted EFT images from a circular camera trajectory to initialize a rough volume. During the next 2,000 steps, we perform diffusion distillation. Reconstructing a single instance takes roughly an hour on an A5000 gpu.


\subsection{Reconstruction on Real Images}

\vspace{-2mm}
\paragraph{Core Subset: 2-view.} We show 2-view category-specific reconstruction results for the 10 \emph{core subset} categories. We evaluate metrics on the first 10 scenes of each category. For each scene, we load 32 linearly spaced views, from which we randomly sample two input views and evaluate on the remaining 30 unseen views. The input and evaluation views are held constant across methods. We report category-specific PSNR and LPIPS in \tableref{co3d-10d}. We show qualitative comparisons in  \figref{comparison-large}.

SparseFusion outperforms all other methods in LPIPS, only losing out in PSNR for 3 categories.  Despite PSNR favoring mean predicting methods, SparseFusion achieves higher PSNR in 7 categories. The strong performance of SparseFusion is reflected in the qualitative comparison. Existing methods either predict a blurry view for unseen regions or a perceptually reasonable view that disregards 3D consistency. SparseFusion predicts views that are both perceptually reasonable and geometrically consistent.

\vspace{-2mm}
\paragraph{Core Subset: Varying Views.}
We examine performance of the different methods as we increase the number of input views. As the number of input views increases, more regions are observed, giving an advantage to methods that explicitly use feature re-projection. We evaluate 2, 3, and 6 view reconstruction on the \emph{core subset} categories and show PSNR, SSIM, and LPIPS in Table \ref{tab:co3d-10}.

We see feature re-projection methods improve drastically with more input views as the need for hallucination of unseen regions decreases. EFT outperforms SparseFusion in PSNR for the 3-view and 6-view settings. However, SparseFusion remains competitive in PSNR while being better in LPIPS. SSIM results further underscore the advantage of SparseFusion with sparse (2, 3) input views. Moreover, SparseFusion outperforms all current state-of-the-art methods in all three metrics for 2, 3, and 6 view reconstruction.

\begin{figure}[t]
\centering 
\vspace{-0.5mm}
\includegraphics[width=1.0\linewidth]{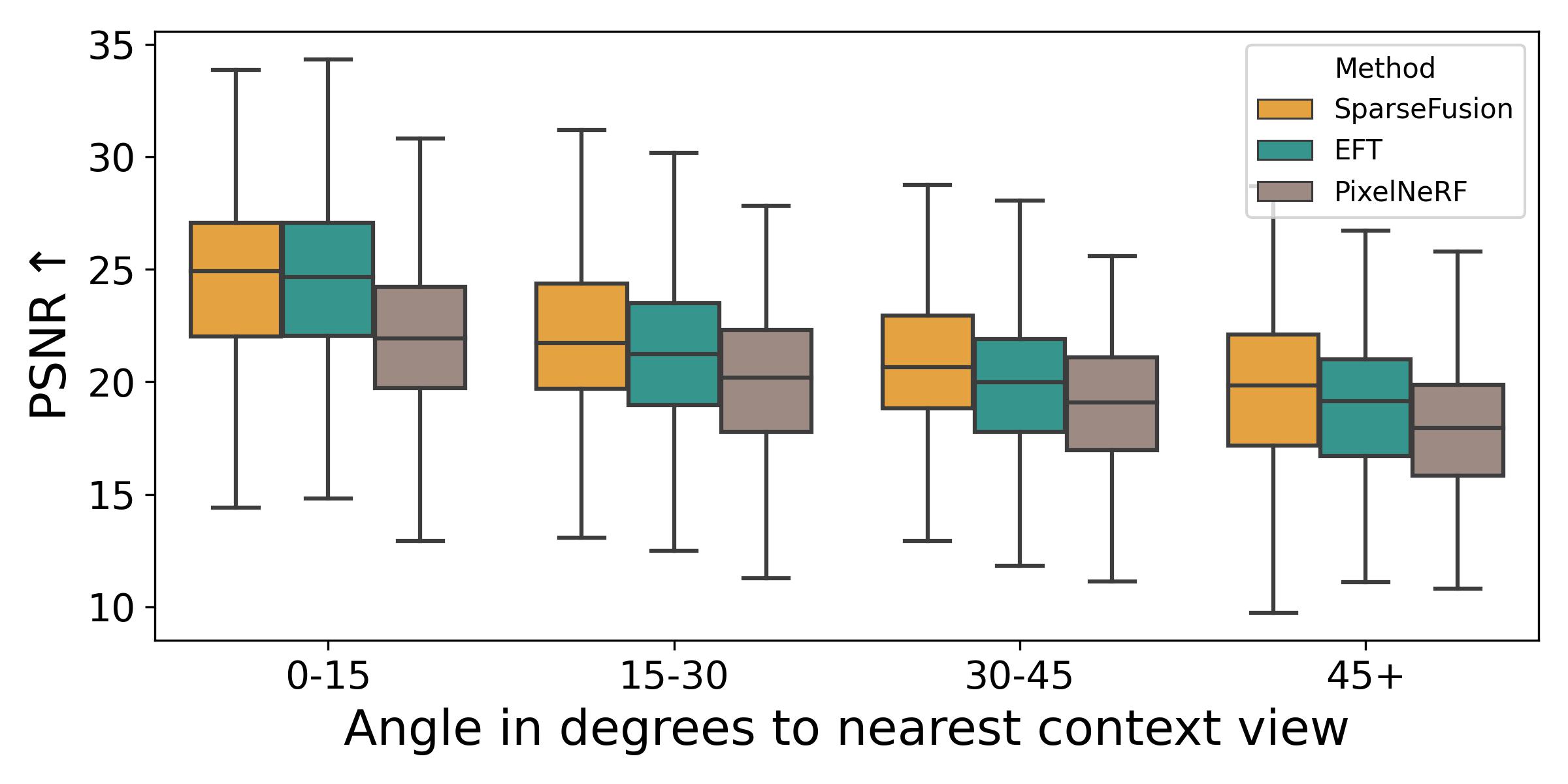}
\includegraphics[width=1.0\linewidth]{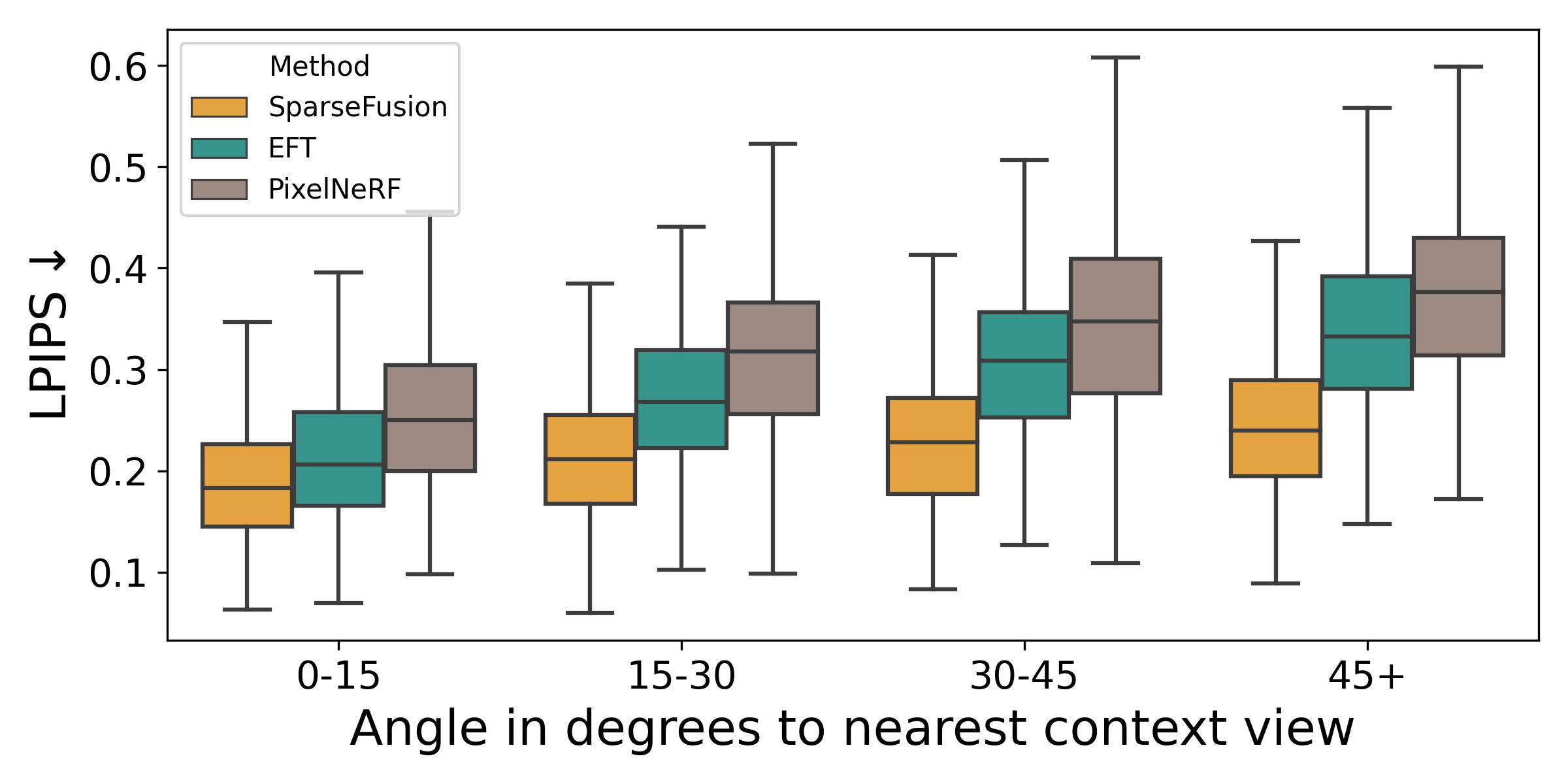}
\vspace{-1.5em}
\captionof{figure}{
    \textbf{Metrics Binned by Viewpoint Change.} We show metrics binned by the angle of query camera to the nearest context view. Results are aggregated from  \tableref{co3d-10d}.
} 
\figlabel{view-angles}
\vspace{-1em}
\end{figure}

\begin{figure*}[t!]
    \centering
    \includegraphics[width=1.0\linewidth]{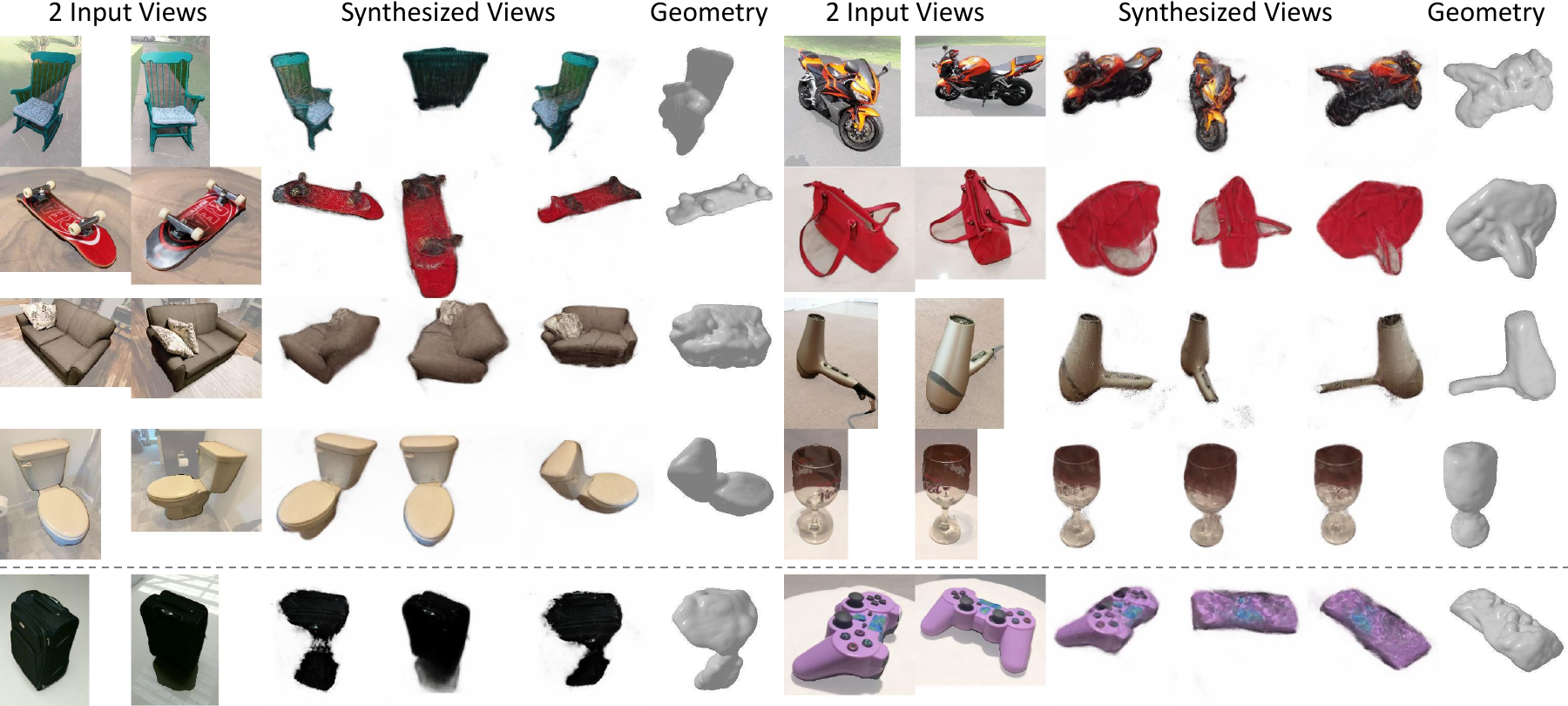}
    \caption{
    \textbf{Reconstruction on Diverse Categories.} We show SparseFusion reconstructions on a subset of the 51 CO3D categories. We also show a couple of failure modes on the last row. Please see project page for more samples and 360-degree visualizations.}
    \vspace \figmargin
    \label{fig:diverse}
\end{figure*} 
\begin{table}
\footnotesize

\begin{center}

\caption{\textbf{The Importance of Mode Seeking.} We show metrics when EFT and VLDM are naively used to optimize Instant NGP \cite{mueller2022instant} in a mean seeking behavior, versus the mode seeking optimization in SparseFusion. We average across 10 scenes of hydrants with 2 input views.}
\vspace{-.5em}
\tablelabel{modeseeking}

\begin{tabular}{llccc}
\toprule
Backbone & Method & PSNR $\shortuparrow$ & SSIM $\shortuparrow$ & LPIPS $\shortdownarrow$ \\
\midrule
\multirow{2}{*}{EFT}
    & base                  & 21.58 & 0.732 & 0.224 \\
    & base w/ INGP          & 21.57 & 0.780 & 0.219 \\
\midrule
\multirow{3}{*}{VLDM}
    & base                  & 20.05 & 0.776 & 0.178 \\
    & base w/ INGP          & 20.61 & 0.753 & 0.230 \\
    & SparseFusion          & \textbf{22.35} & \textbf{0.817} & \textbf{0.153} \\
\bottomrule
\vspace{-3em}
\end{tabular}
\end{center}
\end{table}

\vspace{-2mm}
\paragraph{All Categories: 2-views.}
We compare against NerFormer and ViewFormer across all 51 categories to demonstrate SparseFusion's performance on diverse categories. We evaluate with 2 random input views on the first 5 scenes of each category for all 51 categories and report the averaged metrics in Table \ref{tab:co3d-50}. While EFT edges out in PSNR, SparseFusion achieves better SSIM and LPIPS. Existing methods, NerFormer and ViewFormer perform significantly worse. We show qualitative results of SparseFusion on diverse categories in Figure \ref{fig:diverse}\, where, in addition to 3  synthesized novel views, we also visualize the underlying geometry by extracting an iso-surface via marching cubes.


\begin{table}
\footnotesize

\begin{center}
\vspace{.5em}
\caption{\textbf{Diffusion Distillation Setup.} We show that a combination of multi-step prediction and perceptual loss strikes a balance between all three metrics. (hydrant, 10 scenes, 2 input views)}
\vspace{-1em}
\tablelabel{supp-ablation}
\resizebox{\linewidth}{!}{
\begin{tabular}{lllccc}
\toprule
Loss Space & Denoising Steps & Perceptual Loss & PSNR $\shortuparrow$ & SSIM $\shortuparrow$ &LPIPS $\shortdownarrow$ \\
\midrule
\multirow{4}{*}{Latent}
    & \multirow{2}{*}{Single}
        & No                  & 22.25 & 0.720 & 0.211 \\
        & & Yes                & 22.15 & 0.770 & 0.187 \\ \cmidrule{2-6}
    & \multirow{2}{*}{Multiple}
        & No                  & 21.92 & 0.744 & 0.211 \\
        & & Yes                & 22.03 & 0.781 & 0.170 \\
\midrule
\multirow{4}{*}{Pixel}
    & \multirow{2}{*}{Single}
        & No                  & 22.13 & 0.792 & 0.208 \\
        & & Yes                & \textbf{22.49} & \textbf{0.826} & 0.169 \\ \cmidrule{2-6}
    & \multirow{2}{*}{Multiple}
        & No                  & 22.36 & 0.797 & 0.200 \\
        & & Yes                & 22.35 & 0.817 & \textbf{0.153} \\
\bottomrule
\vspace{-3em}
\end{tabular}
}
\vspace{-1em}
\end{center}
\end{table}
\begin{figure}[t!]
\centering 
\vspace{.5em}
\includegraphics[width=0.9\linewidth]{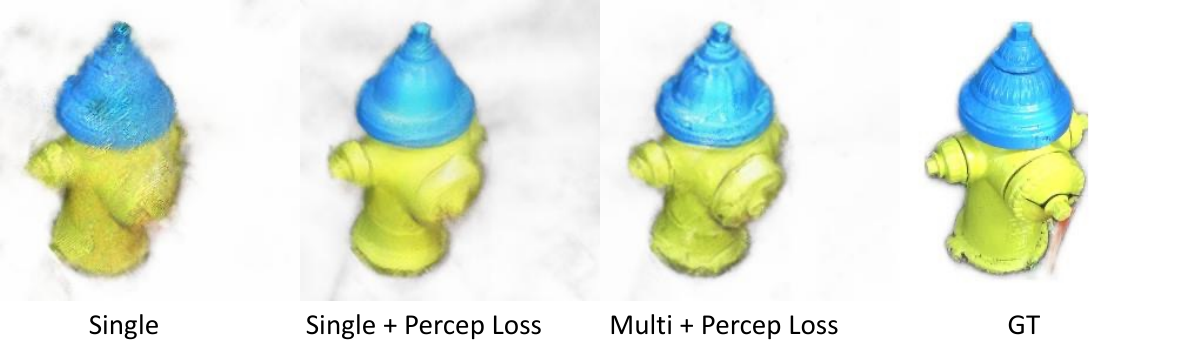}
\vspace{-.5em}
\captionof{figure}{
    \textbf{Qualitative Results with Pixel Space Loss.} Using multi-step denoising and perceptual loss achieves more realistic results. 
} 
\vspace{-2em}
\figlabel{supp-ablation-qual}
\end{figure}

\vspace{-4mm}
\paragraph{Failure Modes.} We show failure modes on the bottom row of Figure \ref{fig:diverse}. On the bottom left, SparseFusion fails to reconstruct a good geometry for the black suitcase. As  Instant NGP is trained to output a default black color for the background,  the neural representation sometimes fails to disambiguate black foreground from black background. On the bottom right, we see SparseFusion propagating a dataset bias for the category, remote. Since most remote images are TV remotes, SparseFusion attempts to make the video game controller a TV remote.

\subsection{Additional Analysis}

\paragraph{Performance Binned by Viewpoint Changes.}
We investigate the relationship between magnitude of viewpoint change and reconstruction performance. We analyze SparseFusion, EFT, and PixelNeRF results on the \emph{core subset} and visualize PSNR and LPIPS binned by angle in degrees to the nearest context view in \figref{view-angles}. We show that for small viewpoint changes, SparseFusion performs better in LPIPS and competitively in PSNR against EFT. As viewpoint change increases, feature re-projection methods fall off quite fast while SparseFusion remains more robust and performs relatively better.  

\vspace{-2mm}
\paragraph{Importance of Mode Seeking.}
We compare the diffusion distillation formulation against a naive method to  obtain a neural representation  given a view synthesis method (VLDM or EFT). Concretely, we obtain several rendered samples $(\{\hat{I}, \hat{\bm{\pi}}\})$ from the base view synthesis method given the context views $C$, and simply train an INGP to fit a 3D representation to these.

We  present the results in \tableref{modeseeking},
and see no significant change when we fit INGP to EFT renderings because EFT predicts consistent mean outputs. However, when we fit INGP to VLDM predictions, we see that perceptual quality decreases. We show a qualitative example in \figref{mode} and also illustrate a toy 2D scenario which explains this drop due to mean seeking  where averaging over conflicting samples leads to a poor reconstruction. However, when we optimize INGP using the diffusion distillation objective, all metrics improve, underscoring the importance our proposed of mode seeking optimization. 

\vspace{-2mm}
\paragraph{Ablating Distillation Objective.}
We examine performance across various distillation design choices in \tableref{supp-ablation}. We observe that for all methods, PSNR remains relatively similar. However, computing loss in pixel space and additionally using perceptual loss improves both SSIM and LPIPS. Moreover, the multi-step denoising leads to the best perceptual results. 
While single-step denoising with perceptual loss achieves better PSNR and SSIM by a small margin, qualitative results in \figref{supp-ablation-qual} show that the predicted texture is smooth and unrealistic.
\section{Discussion}
\seclabel{conclusion}
We presented an approach for inferring 3D neural representations from sparse-view observations. Unlike prior methods that struggled to deal with uncertainty, our approach allowed predicting 3D-consistent representations with plausible and realistic outputs even in unobserved regions. While we believe our work represents a significant step forward in recovering detailed 3D from casually captured images, a few challenges still remain. 
A key limitation of our work (as well as prior methods) is the reliance on known (relative) camera poses across the observations, and while there have been recent promising advances~\cite{zhang2022relpose, rockwell2022}, this remains a challenging task in general.
Additionally, our approach requires optimizing instance-specific neural fields and is computationally expensive. Finally, while our work introduced the view-conditioned diffusion distillation in context of sparse-view reconstruction, we believe even single-view 3D prediction approaches can benefit from leveraging similar objectives.


\section{Ethics and Broader Impact}
\seclabel{discussion}
Compared to existing novel view synthesis methods, SparseFusion is more computationally expensive. This poses a hardware limitation for potential downstream tasks and may also increase carbon emissions. Additionally, SparseFusion relies on view-conditioned latent diffusion models (VLDM), which are trained on multi-view datasets. VLDMs are good at representing their training data, potentially learning harmful biases that will propagate to reconstructed 3D scenes. While our current use case for reconstructing static objects from CO3D categories does not present ethical concerns, adapting SparseFusion to humans or animals requires more thorough examination of bias present in the training data.  
\section*{Acknowledgements}
\seclabel{acknowledgements}

We thank Naveen Venkat, Mayank Agarwal, Jeff Tan, Paritosh Mittal, Yen-Chi Cheng, and Nikolaos Gkanatsios for helpful discussions and feedback. We also thank David Novotny and Jonáš Kulhánek for sharing pretrained models for NerFormer and ViewFormer, respectively. This material is based upon work supported by the National Science Foundation Graduate Research Fellowship under Grant No. (DGE1745016, DGE2140739).

{\small
\bibliographystyle{ieee_fullname}
\bibliography{main}
}

\clearpage
\appendix
\renewcommand{\thesection}{\Alph{section}}
\section*{Appendix}

\section{Extended Background: Denoising Diffusion}
\seclabel{extended-background}

Denoising diffusion probabilistic models \cite{ho2020denoising} approximate a distribution $p(\bm{x})$ over real data by reversing a Markov chain of diffusion steps, starting from Gaussian noise at $\bm{x}_T$ to a realistic image at $\hat{\bm{x}}_0$. See \cite{ho2020denoising} for details.

\paragraph{Forward Process.}
The forward diffusion process, which incrementally adds noise to a real image $\bm{x}_0$ until the image becomes Gaussian noise $\bm{x}_T$, is defined in \eqref{supp-diffusion-forward}. Forward variance $\beta$ is usually defined by a fixed schedule.

\begin{equation}
\eqlabel{supp-diffusion-forward}
q(\bm{x}_t|\bm{x}_{t-1}) = \mathcal{N}(\sqrt{1-\beta_t}\bm{x}_{t-1}, \beta_t\bm{I})
\end{equation}

\paragraph{Reverse Process.}
The reverse diffusion process reverses the noise added in the forward process, effectively denoising a noisy image. When we generate a sample from a diffusion model, we apply the reverse process $T$ times from $t=T$ to $t=1$. The reverse process is defined in \eqref{supp-diffusion-reverse}, where posterior mean $\mu_{\phi}(\bm{x}_t, t)$ is predicted from a network and posterior variance $\sigma^2$ follows a fixed schedule (though other works such as  \cite{song2020denoising} also learn  $\sigma^2$ with a network). 

\begin{equation}
\eqlabel{supp-diffusion-reverse}
p(\bm{x}_{t-1}|\bm{x}_t) = \mathcal{N}(\mu_{\phi}(\bm{x}_t, t), \sigma^2\bm{I})
\end{equation}

\paragraph{Posterior Mean.} Prior works \cite{ho2020denoising, song2020denoising} have found that parameterizing the neural network to predict $\bm{\epsilon}$ instead of $\bm{x}_{t-1}$ or $\bm{x}_0$ works better in practice. We write posterior mean in terms of $\bm{\epsilon}$ in \eqref{supp-mu} where $\alpha_t = 1 - \beta_t$ and $\bar{\alpha}_t = \Pi_{s=1}^t \alpha_s $.
\begin{equation} \eqlabel{supp-mu}
\mu_\phi(\bm{x}_t, t) = \frac{1}{\sqrt{\alpha_t}}(\bm{x}_t - \frac{\beta_t}{\sqrt{1-\bar{\alpha_t}}} \bm{\epsilon}_\phi(\bm{x}_t, t))
\end{equation}

As mentioned in the main text, this parametrization leads to a training framework where one adds (time-dependent) noise to a data point $\bm{x}_0$, and then trains the network $\bm{\epsilon}_\phi$ to predict this noise given the noisy data point $\bm{x}_t$.
\begin{equation} \eqlabel{diffusion}
\begin{split}
\mathcal{L}_{DM} = \mathbb{E}_{{\bm{x}_0}, \bm{\epsilon}, t}\left[w_t~||\bm{\epsilon} - \bm{\epsilon}_\phi(\bm{x}_t, t)||^2\right]
\\ \text{where}~\bm{x}_t = \sqrt{\bar{\alpha_t}} \bm{x}_0 + \sqrt{1-\bar{\alpha_t}} \bm{\epsilon}; ~~ \bm{\epsilon} \sim \mathcal{N}(0,1)
\end{split}
\end{equation}

In this work, we use conditional diffusion models to infer distributions of the form $p(\bm{x}|\bm{y})$ by additionally using $\bm{y}$ as an input for the noise prediction network $\bm{\epsilon}_\phi(\bm{x},\bm{y}, t)$. 
\section{Implementation Details}
\seclabel{implementation}
We provide detailed implementation and training details for all components of SparseFusion.

\subsection{Epipolar Feature Transformer}
\seclabel{impl-eft}

\paragraph{Overview.}
Epipolar feature transformer is a feed-forward network that first gathers features along the epipolar lines of input images before aggregating them through a series of transformers. EFT is inspired by the GPNR approach by Suhail \etal \cite{suhail2022generalizable}, but we modify the feature extractor backbone to better suit the sparse-view setup and additionally use epipolar features for conditional diffusion. We describe our implementation below.

\parnobf{Notation:} Let $g_\psi$ be the RGB branch and $h_\psi$ be the feature branch.
\parnobf{Inputs:} $C \equiv {(\bm{x}_m, \bm{\pi}_m})$, a set of input images with known camera poses and a query pose $\bm{\pi}$ -- note that the poses are w.r.t. an arbitrary world-coordinate system and we only use their relative configuration.
\parnobf{Outputs:} an RGB image $\bm{x}$ and a feature grid $\bm{y}$ corresponding to the query viewpoint $\bm{\pi}$.

\paragraph{Feature Extractor Backbone.}
Given input views $C \equiv {(\bm{x}_m, \bm{\pi}_m})$ where $\bm{x}_m$ is the $m^{th}$ masked (black background) input image of shape (256, 256, 3). We use ResNet18 \cite{he2016deep} as our backbone to extract pixel-aligned features by concatenating intermediate features from the first 4 layer groups of ResNet18, using bilinear upsampling to ensure all features are 128 by 128. For each image $\bm{x}_m$, we arrive at a feature grid of shape (128, 128, 512).

\paragraph{Epipolar Points Projector.}
Given a query camera $\bm{\pi}$, each pixel in its image plane corresponds to some ray. Our Epipolar Transformer seeks to infer per-pixel colors or features, and does so by processing each ray using the multi-view projections of points along it. For each ray $\bm{r}$ (parameterized by its origin and direction), we project 20 points along the ray direction with depth values linearly spaced between $z\_near$ and $z\_far$. We set $z\_near$ to $s-5$ and $z\_far$ to $s+5$ where $s$ is the average distance from scene cameras to origin computed per scene. The 20 points, with shape (20, 3), are then projected into the screen space of each of the $m$ input cameras, giving us epipolar points with shape (M, 20, 2). We use bilinear sampling to sample image features at the epipolar points, giving us combined epipolar features of shape (M, 20, 512) per ray. This becomes the input to our epipolar feature transformer. 

\paragraph{Epipolar Feature Transformer.}
\begin{figure}[h!]
    \centering
    \includegraphics[width=1.0\linewidth]{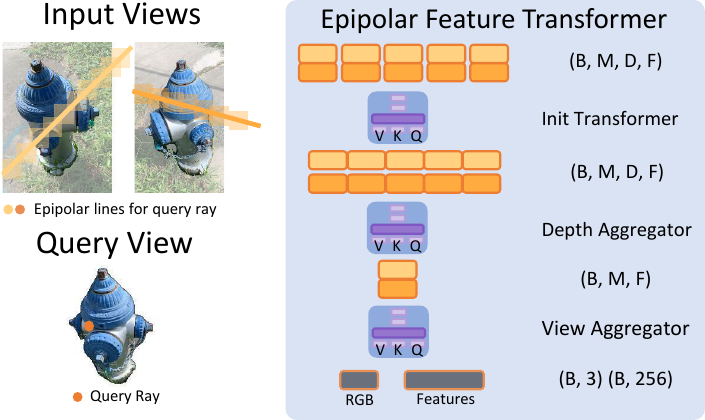}
    \caption{
    \textbf{Epipolar Feature Transformer} We show a diagram of EFT. This module processes each query ray independently, using a transformer to aggregate the projected features across views and across possible depths. For each ray, the output is a predicted RGB color (used as a baseline prediction method), and a pixel-aligned feature (used as conditioning in the diffusion model).}
    \vspace \figmargin
    \figlabel{supp-eft}
\end{figure} 
EFT aggregates the epipolar features from a single ray with a series of three transformers to predict an RGB pixel color and  a 256-dimension feature. We visualize the EFT in \figref{supp-eft}. We show details of the transformers in \tableref{impl-gpnr}. All transformer encoders have hidden and output dimensions of 256. Both the depth aggregator and view aggregator transformers are followed by a weighted average operation, where the output features from the transformers are multiplied by a weight, which sums to 1 along the sequence length dimension. The relative weights are predicted by a linear layer before passing through softmax. This effectively performs weighted averaging along the sequence dimension. 

\begin{table}[h]
\footnotesize

\begin{center}

\caption{\textbf{EFT Configuration.} We use default PyTorch hyperparameters for each layer. B is number of rays. M is the number of input views. D is the number of epipolar feature samples along the ray. D is 20.}
\tablelabel{impl-gpnr}

\resizebox{\linewidth}{!}{
\begin{tabular}{llcc}
\toprule
Transformer & Layers & Sequence Dims / Dim & Output Shape \\
\midrule
\multirow{1}{*}{Init Transformer}
    & Transformer Encoder x4                  & M  & (B, M, D, 256) \\
\midrule
\multirow{3}{*}{Depth Aggregator}
    & Transformer Encoder x4                   & D & (B, M, D, 256) \\
    & Linear + Softmax                                  &  D   & (B, M, D, 1) \\
    & Weighted Average                 &    & (B, M, 256) \\
\midrule
\multirow{3}{*}{View Aggregator}
    & Transformer Encoder x4                    & M & (B, M, 256) \\
    & Linear + Softmax                               & M   & (B, M, 1) \\
    & Weighted Average                  &   & (B, 256) \\
\midrule
\multirow{1}{*}{Color Branch}
    & Linear                   & & (B, 3) \\
\bottomrule
\vspace{-3em}
\end{tabular}
}
\end{center}
\end{table}

The inputs to the transformer are the sampled features concatenated with additional ray and depth encodings. Given a point along the query ray $\bm{r}_q$ at depth $d$, we denote by $\bm{p}_{md}$ its projection in the $m^{th}$ context view. In addition to the pixel-aligned feature $\bm{f}_{md}$ (described in previous paragraph), we also concatenate encodings of the query ray $\bm{r}_q$, the depth $\bm{d}$, and the ray $\bm{r}_{md}$ connecting the $m^{th}$ camera center  to the 3D point. We use plucker coordinates to represent each ray, and compute harmonic embeddings for each to $(\bm{r}_q, \bm{r}_{md}, \bm{d})$ (using 6 harmonic functions) before concatenating them with $\bm{f}_{md}$ to form the input tokens to the transformer.

\paragraph{Training Procedure.}
We can train the color branch of EFT as a standalone novel view synthesis baseline. In our work, EFT is jointly trained with VLDM. Please see supplementary \secref{impl-vldm} for details.

\subsection{View-conditioned Diffusion Model}
\seclabel{impl-vldm}

\paragraph{Overview.} View-conditioned diffusion model is a latent diffusion model that conditions on a pixel-aligned feature grid $\bm{y}$. 

\parnobf{Notation:} Let $\epsilon_\phi$ be the denoising UNet, $\mathcal{E}$ be the VAE encoder, and $\mathcal{D}$ be the VAE decoder.

\paragraph{VAE.}
We use the VAE from Stable Diffusion \cite{rombach2022high}. We use the provided v1-3 weights and keep the VAE frozen for all experiments. We use (256, 256, 3) RGB images as input, and the VAE encodes them into latents of shape (32, 32, 4). We refer readers to \cite{rombach2022high} for more details. 

\paragraph{Denoising UNet.}
Our 400M parameter UNet roughly follows \cite{saharia2022photorealistic}. We construct our UNet using code from \cite{imagen-pytorch} with the parameters in \tableref{impl-unet-overview}. 
\begin{table}[h!]
\footnotesize

\begin{center}

\caption{\textbf{UNet Parameters.} We provide parameters for our UNet.}
\tablelabel{impl-unet-overview}

\begin{tabular}{ll}
\toprule
Parameter &  Value\\
\hline
channels & 4\\
dim & 256\\
dim\_mults & (1,2,4,4)\\
num\_resnet\_blocks & (2,2,2,2) \\
layer\_attns & (False, False, False, True)\\
cond\_images\_channels & 256\\
\bottomrule
\vspace{-3em}
\end{tabular}
\end{center}
\end{table}

The UNet comprises of 4 down-sampling blocks, a middle block, and 4 up-sampling blocks. We show the input and output shape for the modules of the UNet in \tableref{impl-unet}. We refer readers to \cite{imagen-pytorch} for UNet details. We disable all text conditioning and cross attention mechanisms; instead, we concatenate EFT features, $\bm{y}$, with image latents, $\bm{z}_t$. These EFT features are computed for the of $32 \times 32$ rays corresponding to the patch centers.

\begin{table}[h]
\footnotesize

\begin{center}

\caption{\textbf{UNet Blocks.} We outline the modules in our denoising UNet.}
\tablelabel{impl-unet}

\begin{tabular}{llcc}
\toprule
Modules & Block & Output Shape \\
\midrule
\multirow{1}{*}{Input}
    &                            & (B, 260, 32, 32) \\
\midrule
\multirow{1}{*}{Init. Conv}
    & InitBlock                     & (B, 256, 32, 32) \\
\midrule
\multirow{1}{*}{Down 1}
    & DownBlock                     & (B, 256, 16, 16) \\
\midrule
\multirow{1}{*}{Down 2}
    & DownBlock                    & (B, 512, 8, 8) \\
\midrule
\multirow{1}{*}{Down 3}
    & DownBlock                   & (B, 1024, 4, 4) \\
\midrule
\multirow{2}{*}{Down 4}
    & DownBlock                    & (B, 1024, 4, 4)  \\
    & Self-attention                    & (B, 1024, 4, 4)  \\
\midrule
\multirow{1}{*}{Middle}
    & MiddleBlock                     & (B, 1024, 4, 4) \\
\midrule
\multirow{2}{*}{Up 1}
    & UpBlock                     & (B, 1024, 8, 8) \\
    & Self-attention                     & (B, 1024, 8, 8) \\
\midrule
\multirow{1}{*}{Up 2}
    & UpBlock                    & (B, 512, 16, 16) \\
\midrule
\multirow{1}{*}{Up 3}
    & UpBlock                    & (B, 256, 32, 32) \\
\midrule
\multirow{1}{*}{Up 4}
    & UpBlock                     & (B, 256, 32, 32) \\
\midrule
\multirow{1}{*}{Final Conv.}
    & Conv2D                     & (B, 4, 32, 32) \\
\bottomrule
\vspace{-3em}
\end{tabular}

\end{center}
\end{table}

\paragraph{Training Procedure.}
We train with batch size of 2, randomly chosen number of input views between 2-5, and learning rate of 5e-5 using Adam optimizer with default hyperparameters for 100K steps. We optimize both the UNet weights and also the EFT weights. We optimize the UNet and feature branch of EFT with the simplified variational lower bound \cite{ho2020denoising}. We optimize the color branch of EFT with pixel-wise reconstruction loss.

\subsection{Diffusion Distillation}
\seclabel{impl-distillation}

\paragraph{Overview.} We optimize a 3D neural scene representation, Instant NGP \cite{mueller2022instant, torch-ngp}, with our VLDM. 

\parnobf{Notation:} Let $f_\theta$ be the volumetric Instant NGP renderer, $p_\phi(\bm{z}_{0:\mathcal{T}}| \bm{\pi}, C)$ be the multi-step denoising process that estimates $\hat{\bm{z}}_0$. Let $\Pi$ be an instance-specific camera distribution.

\paragraph{Instant NGP.} We use the PyTorch Instant NGP implementation from \cite{torch-ngp}. We set scene bounds to 4 with desired hashgrid resolution of 8,192. We use a small 3 layer MLP with hidden dimension of 64 to predict RGB and density. We do not use view direction as input. 

\paragraph{Camera Distribution.}
Given a set of input cameras $C_I \equiv {(\bm{\pi}_m})$ and a query camera $\bm{\pi}_q$, we first find the look-at point $P_{at}$ by finding the nearest point to all $m+1$ rays originating from camera centers. Then, we fit a circle $O$ in 3D space with center being the mean of all camera centers. Let the normal of circle $O$ be $\bm{n}$. To sample a camera, we first sample a point $P_i$ on $O$ and jitter the angle between $\overline{P_{at}P_i}$ and $\bm{n}$ by $\mathcal{N}(0, 0.17)$ radians to get jittered point $P_{i}^{'}$. We then construct a camera $\bm{\pi}$ with center $P_{i}^{'}$ looking at $P_{at}$.

\paragraph{Multi-step Diffusion Guidance.}
Given a rendered image $\bm{x}_0$, we encode it to obtain $\bm{z}_0$. Then, we uniformly sample $t \sim (0,T]$ and construct a noisy image latent $\bm{z}_t$. We perform multi-step denoising to obtain $\hat{\bm{z}}_0$ by iteratively sampling $\hat{\bm{z}}_{t_{k-1}} \sim p_\phi(\bm{z}_{t_{k-1}}|\hat{\bm{z}}_{t_k}, y)$ on an interval of time steps $\mathcal{T} = (t_1, ..., t_k, t)$ using a linear multi-step method \cite{liu2022pseudo}. We construct $\mathcal{T}$ by linearly spacing $k+1$ time steps between $(0, t]$. We define $k$ with a simple scheduler: 

\begin{equation}
\eqlabel{k-scheduler}
k + 1 = 
\left\{
    \begin{array}{lr}
        \frac{100t}{T}, & \text{if } t \leq \frac{T}{2}\\
        50, & \text{if } t > \frac{T}{2}
    \end{array}
\right\}
\end{equation}

Finally, given $\hat{\bm{z}}_0$, we get the predicted image $\hat{\bm{x}}_0 = \mathcal{D}(\hat{\bm{z}}_0)$. We do not compute gradients through multi-step diffusion and treat $\hat{\bm{x}}_0$ as a detached tensor.  

\paragraph{Distillation Details.}
We perform 3,000 steps of distillation, optimizing weights of the MLP $\theta$ with Adam optimizer and learning rate 5e-4. During each step of diffusion distillation, we sample $\bm{\pi} \sim \Pi$ and render an image $\bm{x}_0 = f_\theta(\bm{\pi})$. For the first 1,000 steps, we compute rendering loss between $f_\theta(\bm{\pi})$ and $g_\psi(\bm{\pi}|C)$. During the remaining steps, we compute loss between $f_\theta(\bm{\pi})$ and $\hat{\bm{x}}_0$ and use weighting $w_t = 1 - \bar{\alpha}_t$. To avoid out-of-memory error, we render images at reduced resolution (128, 128) and apply bilinear up-sampling before performing multi-step diffusion. In addition, we compute rendering loss between $f_\theta(\bm{\pi}_m)$ and $\bm{x}_m$ on all $m$ input images. Optimizing a single scene takes roughly 1 hour on an A5000 GPU. 

\end{document}